\documentclass{article}

\PassOptionsToPackage{numbers, compress}{natbib}

\usepackage[final]{nips_2018}

\usepackage[utf8]{inputenc} 
\usepackage[T1]{fontenc}    
\usepackage{hyperref}       
\usepackage{url}            
\usepackage{booktabs}       
\usepackage{amsfonts}       
\usepackage{nicefrac}       
\usepackage{microtype}      
\usepackage{amsmath}
\usepackage{graphicx}
\usepackage{color}
\usepackage[tableposition=top]{caption}
\usepackage{subcaption}

\title{Vision-Based Gait Analysis for Senior Care}

\author{
  David Xue\thanks{These authors contributed equally to this work.} , Anin Sayana\footnotemark[1] , Evan Darke\footnotemark[1] , Kelly Shen\footnotemark[1] ,\\ \textbf{Jun-Ting Hsieh, Zelun Luo, Li-Jia Li, N. Lance Downing, } \\
  \textbf{Arnold Milstein, Li Fei-Fei} \\
  Stanford University \\
  \texttt{\{junting, alanzluo\}@stanford.edu}
}

\begin{document}

\maketitle

\begin{abstract}
As the senior population rapidly increases, it is challenging yet crucial to provide effective long-term care for seniors who live at home or in senior care facilities. Smart senior homes, which have gained widespread interest in the healthcare community, have been proposed to improve the well-being of seniors living independently. In particular, non-intrusive, cost-effective sensors placed in these senior homes enable gait characterization, which can provide clinically relevant information including mobility level and early neurodegenerative disease risk. 
In this paper, we present a method to perform gait analysis from a single camera placed within the home. We show that we can accurately calculate various gait parameters, demonstrating the potential for our system to monitor the long-term gait of seniors and thus aid clinicians in understanding a patient's medical profile. 

\end{abstract}

\section{Introduction}

Many countries are currently experiencing a substantial shift in age demographics, with a rising proportion of the geriatric-aged population.
Unfortunately, support for this rapidly growing aging population is becoming more tenuous. It is estimated that the ratio of caregivers to seniors will drop to 4 in 2030, and further fall below 3 in 2050~\cite{redfoot2013aging}. In an effort to support caregivers, there has been increasing awareness around smart senior homes -- technologies that not only enable frail seniors to remain safely independent in a more sustainable fashion, but can also continuously monitor their health status to serve preemptively assess disease risk~\cite{courtney2008privacy,demiris2008senior,luo2018computer}. 

Our goal is to build a vision-based system that automatically analyzes a senior's long-term gait patterns from a single camera placed within the home. Gait analysis is a widely-studied field and has diverse applications in sports, rehabilitation, and health diagnostics~\cite{perry1992gait}. 
Gait speed in particular has been shown to be a simple and effective measure of mobility level~\cite{montero2004gait} for seniors living independently at home or senior care facilities~\cite{studenski2011gait}. Abnormal gait pattern has been shown to be an indicator of diseases such as progressive dementia, residual hemiplegia and Parkinson's disease~\cite{wall1986gait, yogev2007gait}. For instance, previous work assessing spatio-temporal characteristics of patients with Parkinson's disease has revealed significantly higher left-right gait asymmetry, even in early stages~\cite{baltadjieva2006marked,plotnik2005freezing}. 

In cases where a senior has a neurodegenerative disease such as Parkinson's, there is a significant need for constant evaluation of gait-related metrics. Unfortunately, infrequent analysis of neurological metrics over short time periods fails to accurately capture a profile of the patient's health. Reliable characterization of gait can only be done through effective long-term monitoring that captures adequate situational heterogeneity, and allows medical professionals to track changes over time~\cite{muro2014gait}. 

To close this gap, an automated, passive, vision-based system for long-term monitoring of senior's gait can facilitate early diagnosis of neurodegeneration via long-term gait characterization.  In our camera-based system, we demonstrate that we can obtain several important clinical endpoints  from video recordings alone that may potentially indicate onset of Parkinson's, such as gait speed and stride length. Using our method, we can provide long-term gait characteristics, e.g. gait speed across a year, which can be used by doctors to rapidly identify health decline and take immediate, preventative measures to improve patient prognosis. Crucially, this method can be incorporated in any camera-based smart home system, such as in~\cite{luo2018computer}, without introducing additional hardware.

\begin{figure}
  \centering
  \includegraphics[width=0.92\linewidth]{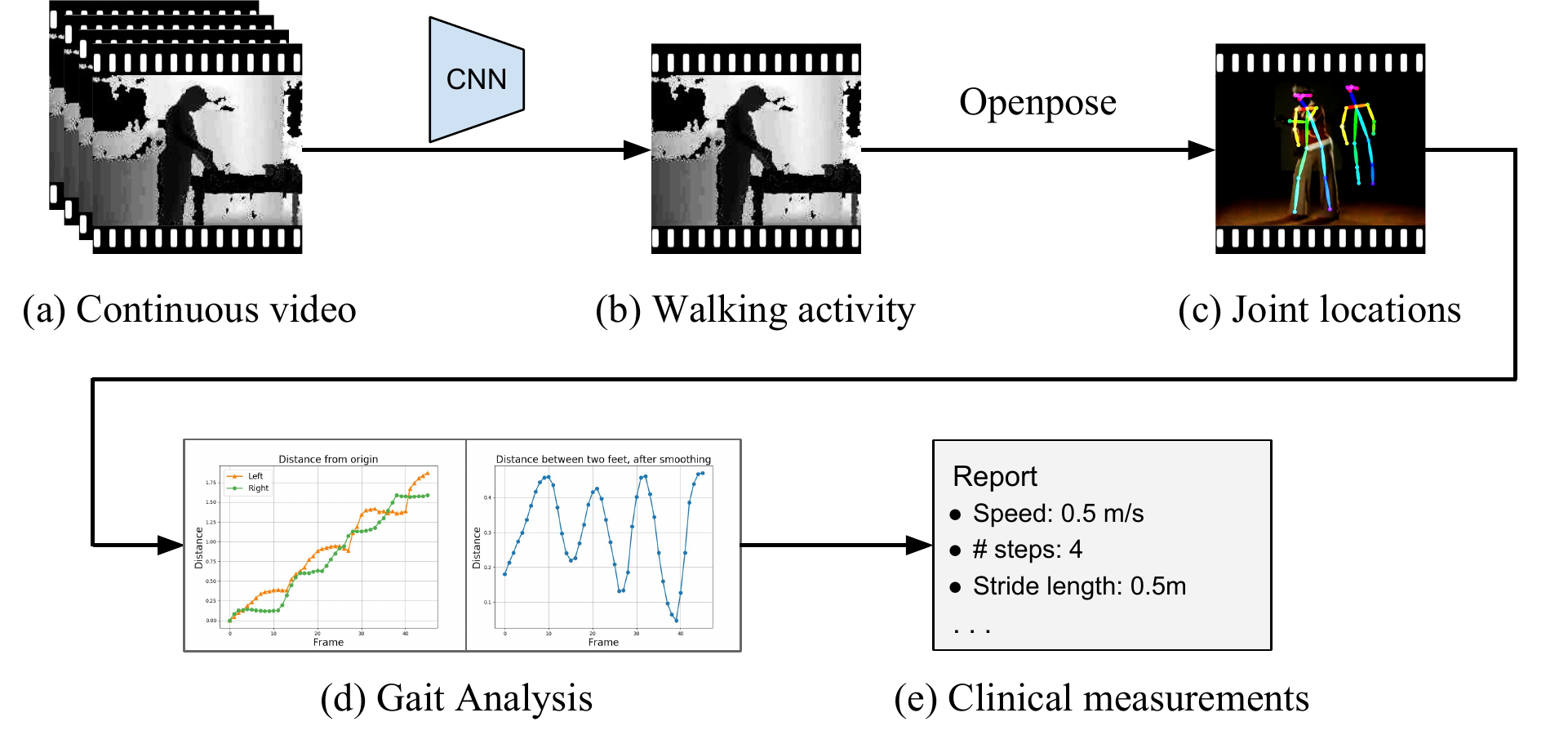}
  \caption{Pipeline of our system. Given a continuous video, we first identify the walking activity using an action detection model. Then, we use Openpose~\cite{cao2017realtime} to extract the 3D joint locations of the senior. Finally, we analyze the gait and calculate important clinical measurements.}
  \label{fig:pull_figure}
\end{figure}

Figure~\ref{fig:pull_figure} shows the pipeline of our system. Given a continuous recording of a senior's daily living, we first identify the walking activity using an action detection model~\cite{luo2018computer}. Then, we extract the skeletal joint locations of the senior with a pose estimation model~\cite{cao2017realtime}. Finally, we run our gait analysis algorithm on the joint locations. We show that we are able to get accurate step detection in 85\% of walking sequences and calculate several medically relevant gait parameters.

\section{Related Work}

\subsection{Smart Senior Homes}

Smart senior homes have started to gain interest in the healthcare community as older adults' desire to live independently at home has grown. 
Several smart home projects~\cite{demiris2008senior,elger1998smartbo} utilize a variety of sensors installed at home to enhance the lifestyle of disabled or elderly people. 
However, most of these approaches require specific home settings and installation of multiple sensors. Recently, \cite{luo2018computer} proposed a system that detects activities of daily living from a single camera, providing a cost-effective way to monitor the long-term health conditions of seniors. Our system follows this principle: we automatically analyze the senior's gait using a single camera placed in the home.

\subsection{Gait Analysis}

Most existing work on gait measurement and analysis can only be done in a short period of time (e.g. when they are being attended in clinic), giving biased evaluations and not capturing the long-term gait~\cite{muro2014gait}. Wearable device-based systems can accurately capture a senior's gait, but they must be worn by the senior. Some previous works proposed low-cost methods that use a mobile phone camera to perform step counting~\cite{aubeck2011camera,ozcan2015robust}, but this method requires the senior to hold the phone facing downward at the feet. Similar to our work, \cite{hagler2010unobtrusive} proposed to install motion sensors at home for continuous assessment, but they only capture walking speed and require several sensors installed on the ceiling. Compared to these methods, our work can analyze gait in the long term from a single camera at home.

\section{Methods}






\subsection{Pose Estimation}


The first step in our pipeline is to extract clips of the senior walking from continuous video recording with a trained action detection model \cite{luo2018computer}.
Once we extract the video frames of the walking activity, we use the Openpose~\cite{cao2017realtime} model to get the skeletal joint locations of the senior in each frame. With a proper transformation using the extrinsic and intrinsic camera matrix, we can calculate the 3D joint locations in the video.
We observe that Openpose is able to robustly capture the person in the view, but may occasionally produce false positives. However, we can easily remove them with temporal consistency or the dimension of the skeleton. For example, if a skeleton in one frame does not appear in the other frames in the video or has unreasonable height or width, then it should be a false positive.


\subsection{Step Detection}

Once we have the sequence of joint locations of $T$ frames, we use the feet locations across time to perform gait analysis. Figure~\ref{fig:algorithm} shows the pipeline of our step detection approach.  Figure~\ref{fig:feet_dist} shows the horizontal displacement from the origin of the left and right foot, $(\ell_t, r_t)|_{t=1}^T$. Next, we calculate the horizontal Euclidean distance between the two feet, $d_t = \|\ell_t - r_t\|$. As shown in Figure~\ref{fig:dist_between_feet_raw}, this forms a periodic pattern across time. When a person swings a foot forward to take a step, the distance between the feet first decreases and then increases. Thus, the local maxima in the curve indicate the points when the two feet are wide apart, which correspond to the steps.

However, due to the gait pattern and the errors in pose estimation, the feet distance $d_t$ may not be smooth, resulting in false local extrema. Thus, we first apply a convolution smoothing with a uniform kernel to get a smoothed curve $\bar{d}_t$. 
Figure~\ref{fig:dist_between_feet_smoothed} shows the curve after smoothing, in which some false extrema are removed while the correct ones are preserved. Finally, shown in Figure~\ref{fig:extrema}, we determine the correct extrema $m_1, n_1, m_2, n_2 ...$, where $m_i$, $n_i$ refer to the maxima and minima, respectively. Since a step occurs when the feet are farthest apart, the maxima $m_i$ indicate when each step happens.


We observe that the above algorithm works well in practice for most video sequence. However, in some extreme cases, the distance between feet curve still contains false extrema even after smoothing. Therefore, we design an algorithm to remove them. First, we calculate the range of the extrema, $r = \max_i m_i - \min_j n_j$. Then, we determine a threshold $\theta = \alpha r$, where $\alpha < 1$. 
For any pair of consecutive extrema, if the difference is smaller than the threshold $\theta$, we remove that pair. 
Figure~\ref{fig:noisy_feet_dist} shows a rare example where the smoothed $\bar{d}_t$ curve still contains false local extrema, and Figure~\ref{fig:fp_removed_extrema} shows that the algorithm can remove the false positives and obtain the correct ones.
Note that in addition, we remove the first extremum, since we don't count the beginning of the video as a step.

\subsection{Gait Analysis}
Following our step detection methodology above, we can calculate the following medically relevant gait parameters: \textit{gait speed, left/right stride length, step length, step width}, and \textit{swing time}, as previously defined in literature~\cite{muro2014gait}.
Each maximum $m_i$ corresponds to each step, and each minimum $n_i$ occurs between steps when the feet are next to each other. 
We determine step length by the distance between two successive placements of the same foot, and swing time by the time difference between two maxima. Step width is defined as the distance between two equivalent points of both feet, which is naturally the minimum points $n_i$. Finally, gait speed is calculated by the sum of the stride length over the total time, to account for the case when the senior does not walk in a straight line.

\newcommand{\w}{0.28}
\begin{figure}[t]
    \centering
    \begin{subfigure}[t]{\w\textwidth}
        \includegraphics[width=\textwidth]{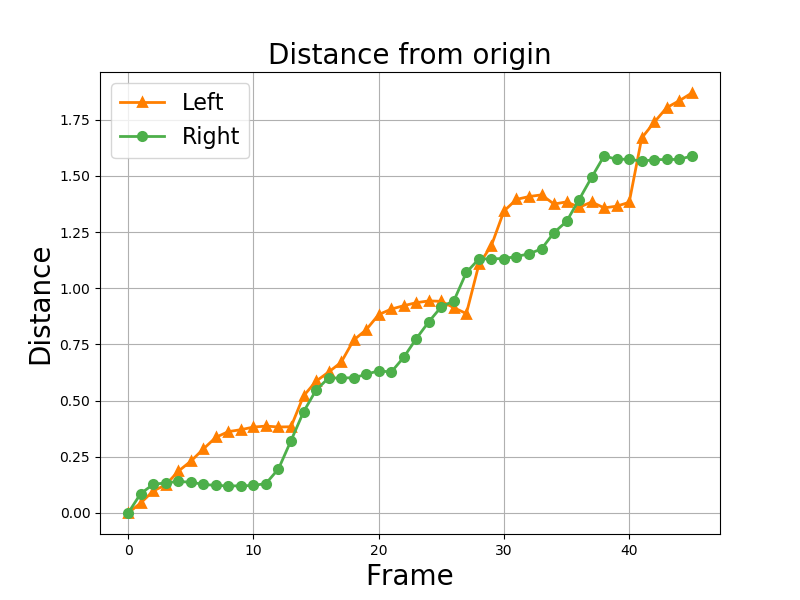}
        \caption{Displacement of each foot from origin.}
        \label{fig:feet_dist}
    \end{subfigure}
    \quad
    \begin{subfigure}[t]{\w\textwidth}
        \includegraphics[width=\textwidth]{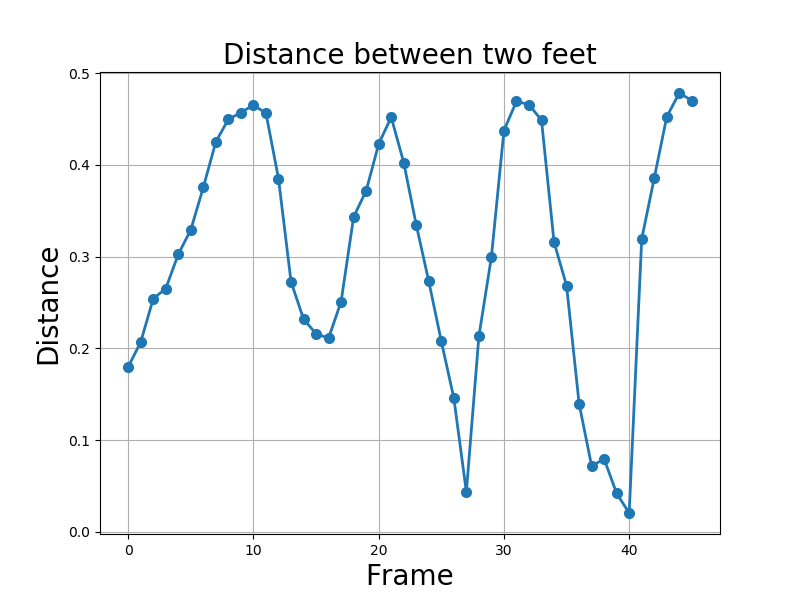}
        \caption{Distance between two feet, $d_t$.}
        \label{fig:dist_between_feet_raw}
    \end{subfigure}
    \quad
    \begin{subfigure}[t]{\w\textwidth}
        \includegraphics[width=\textwidth]{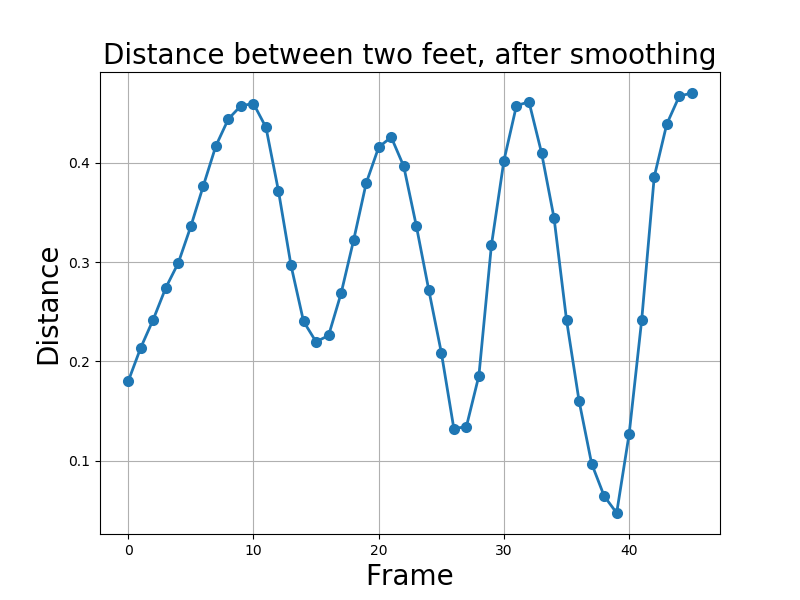}
        \caption{Distance between feet after smoothing, $\bar{d}_t$.}
        \label{fig:dist_between_feet_smoothed}
    \end{subfigure}
    \\
    \begin{subfigure}[t]{\w\textwidth}
        \includegraphics[width=\textwidth]{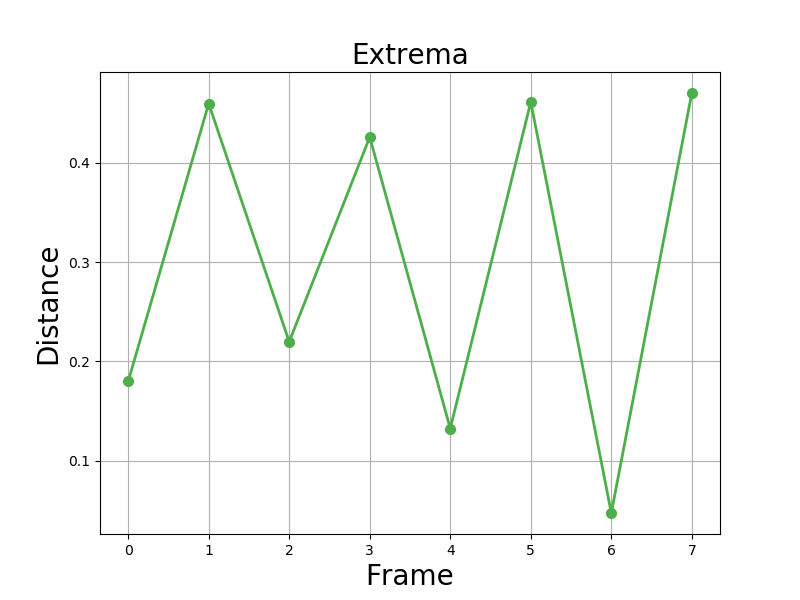}
        \caption{Extrema of the smoothed curve, $m_i$ and $n_i$.}
        \label{fig:extrema}
    \end{subfigure}
    \quad
    \begin{subfigure}[t]{\w\textwidth}
        \includegraphics[width=\textwidth]{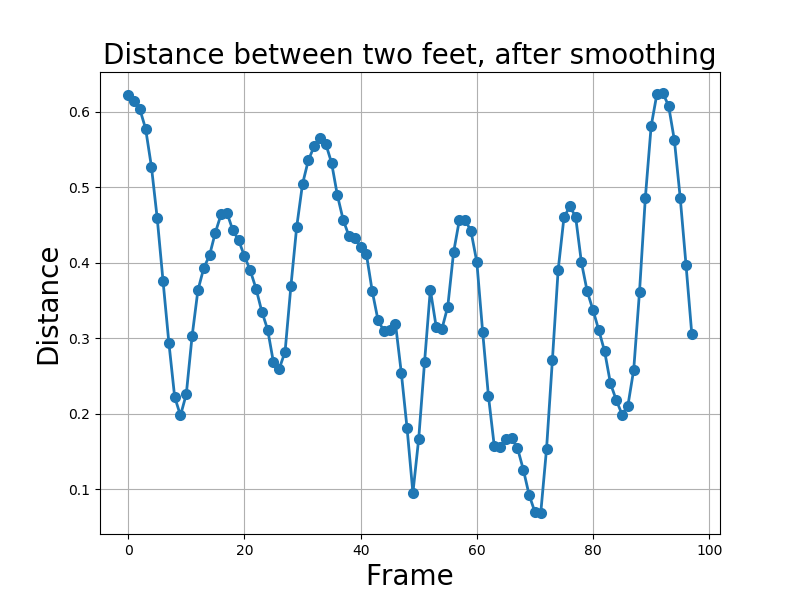}
        \caption{Example of a $\bar{d}_t$ curve with false local extrema.}
        \label{fig:noisy_feet_dist}
    \end{subfigure}
    \quad
    \begin{subfigure}[t]{\w\textwidth}
        \includegraphics[width=\textwidth]{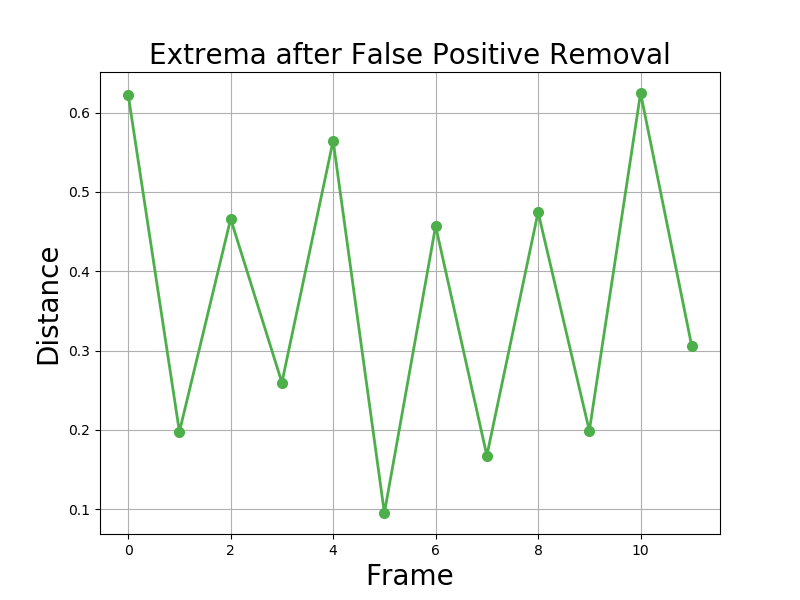}
        \caption{Correct extrema after false positive removal.}
        \label{fig:fp_removed_extrema}
    \end{subfigure}
    \caption{Our gait analysis algorithm:  (a) Extract the 3D feet locations in each frame, (b) calculate the $L2$ distance in the \textit{xy} plane between the two feet, (c) apply smoothing to remove false positives, and finally (d) locate the extrema in the curve.
    (e) In the rare case where the $\bar{d}_t$ curve has false local extrema even after smoothing, (f) we identify the true extrema after false positive removal.}
    \label{fig:algorithm}
\end{figure}

\section{Experiments}


\subsection{Datasets}

\textbf{UTKinect-Action3D}~\cite{xia2012view}. This is a public dataset consisting of videos of 10 human subjects performing different actions in indoor settings. Each video comes with RGB video, depth sequence, and 3D joint locations. We take the videos of the \textit{walking} activity and apply our algorithm.

\textbf{Senior Home Dataset}.
To evaluate our method on videos of seniors, we collected continuous video data recorded from a partner senior care facility. We first extract the video sequences in which the senior is walking, then determine the sequence of joint locations using pose estimation. Finally, we evaluate our algorithm on the dataset. Due to privacy regulations, examples of the data are not shown.

\subsection{Results}

To evaluate our algorithm, we report the ratio of walking sequences where our algorithm correctly identified the correct number of steps. We further measure our accuracy in identifying step occurrence as the average $L1$ distance between our predicted and actual frames for each foot placement. We correctly counted steps in 85\% of examples in the UTKinect-Action3D~\cite{xia2012view} dataset, and detected steps within an average error of $4.4$ frames, or $0.073$ seconds. On the Senior Home dataset, we predicted foot placement with an average error of 0.125 seconds.

We additionally compute medically relevant gait analysis metrics shown in Table~\ref{table:results}. We observe that seniors have a swing time comparable to the average human pace from the UTKinect dataset; however, the senior step length is only 23\% of the UTKinect length, resulting in a proportionally decreased walking speed. We observe that step width appears independent of length and speed. We track both left and right stride in order to detect the presence of asymmetrical stride length, indicative of early onset of neurodegenerative disorders~\cite{fling2018gait}.

\begin{table}
  \caption{Gait analysis results of two examples from the UTKinect-Action3D dataset and our Senior Home dataset. These parameters are defined in~\cite{muro2014gait}.}
  \label{table:results}
  \centering
  \footnotesize
  \begin{tabular}{lccccc}
    \toprule
    Parameters & Speed (m/s) & L/R stride length (m) & Step length (m) & Step width (m) & Swing time (s) \\
    \midrule
    UTKinect & 0.683  & 0.399/0.339 & 0.739 & 0.161 & 0.617  \\
    Senior Home & 0.164 & 0.077/0.092 & 0.169 & 0.131 & 0.625 \\
    \bottomrule
  \end{tabular}
\end{table}

\section{Conclusion}

In summary, we presented an automated system that continuously monitors a senior, detects walking activities, and calculates various gait parameters. Notably, our system only requires a single camera installed at the senior's home, and can be incorporated in any smart system to help medical professionals assess the long-term gait of seniors and understand a patient's medical profile. 

{\small
\bibliographystyle{ieee}
\bibliography{egbib}
}

\end{document}